\documentclass[journal]{IEEEtran}
\usepackage{graphicx}
\usepackage{subcaption}
\usepackage{tabularx}
\usepackage{multirow}
\usepackage{amsmath}
\usepackage{amsfonts}
\usepackage{makecell}
\usepackage{array}
\usepackage{adjustbox}
\usepackage{url}
\usepackage{placeins}
\usepackage[font=footnotesize]{caption}
\begin{document}

%
\title{Generalized Disguise Makeup Presentation Attack Detection Using an Attention-Guided Patch-Based Framework}
%
%
%
\author{
Fateme Taraghi, Atefe Aghaei, and Mohsen Ebrahimi Moghaddam\\
Faculty of Computer science and Engineering\\
Shahid Beheshti University, Tehran, Iran\\
f.taraghi@mail.sbu.ac.ir, a\_aghaei@sbu.ac.ir, m\_moghadam@sbu.ac.ir
}

\maketitle
\begin{abstract}
Despite significant advances in facial recognition systems, they remain vulnerable to face presentation attacks. Among them, disguise makeup attacks are particularly challenging, as they use advanced cosmetics, prosthetic components, and artificial materials to realistically alter facial appearance, often making detection difficult even for humans. Despite their importance, this problem remains underexplored, and publicly available datasets are limited.
To address this, we propose a generalized disguise makeup presentation attack detection framework. The method adopts a two-phase design in which a style-invariant full-face model, trained with metric learning and enhanced by a whitening transformation, extracts region attention scores via Grad-CAM. These scores guide a patch-based phase that performs localized analysis using region-specific subnetworks trained with metric learning for fine-grained discrimination.
We also construct a new, diverse dataset of live and disguise makeup faces collected under real-world conditions, covering variations in subjects, environments, and disguise materials.
Experimental results demonstrate strong generalization across both the collected dataset and SIW-Mv2, achieving 8.97\% ACER and 9.76\% EER on the collected dataset, and 0\% ACER on Obfuscation and Impersonation and 1.34\% on Cosmetics attacks of SIW-Mv2. The proposed method consistently outperforms prior works while maintaining robust performance across other spoof types.
\end{abstract}

\begin{IEEEkeywords}
Face anti-spoofing, disguise makeup detection, presentation attack detection, attention mechanism, patch-based learning, metric learning, style invariance, generalization\end{IEEEkeywords}

\IEEEpeerreviewmaketitle

\section{Introduction}
\IEEEPARstart{I}{n} recent years, face recognition (FR) systems have achieved remarkable progress in automatic identity authentication, driven by advances in computer vision, deep learning, and large-scale datasets \cite{abdullakutty_review_2021,sharma_survey_2023}. These systems are widely used in applications such as border control, remote access, airport security, and device authentication. However, despite their high accuracy, they remain vulnerable to presentation attacks (PAs), which pose serious security risks \cite{abdullakutty_review_2021}.

Presentation attacks attempt to deceive FR systems by presenting fake biometric samples, using methods such as printed photos, replayed videos, 3D masks, or digitally/physically altered faces \cite{sharma_survey_2023}. Among them, makeup-based attacks have become increasingly sophisticated. While makeup is typically used for aesthetic purposes, it can be exploited for identity concealment or impersonation \cite{drozdowski_makeup_2021}. Even conventional cosmetic changes (e.g., eyebrows, lips, skin tone) can significantly alter extracted features, while realistic makeup can closely resemble natural facial characteristics, making detection difficult \cite{kotwal_detection_2020}.

\begin{figure}[!t]
    \centering
    \includegraphics[
    width=0.85\columnwidth,
    height=0.13\textheight,
    keepaspectratio
]{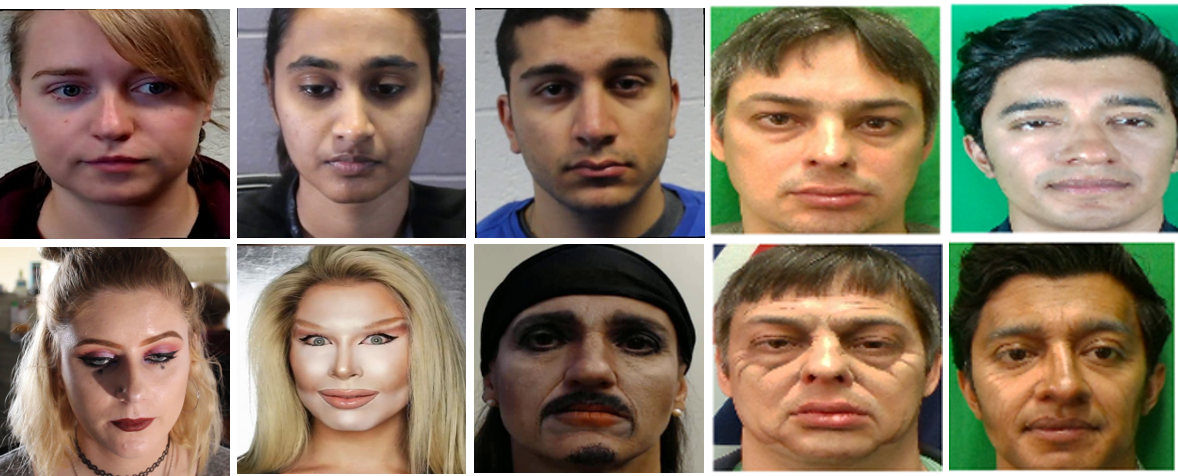}
    \caption{Examples of makeup-based presentation attacks from public datasets. 
    The top row shows live samples, and the bottom row shows spoof samples. 
    Columns 1–3 correspond to cosmetics, impersonation, and obfuscation attacks from the SiW-Mv2 \cite{guo_multi-domain_2022} dataset. 
    Columns 4–5 show aged makeup attacks from the AIM \cite{kotwal_detection_2020} dataset.}
    \label{fig:makeup_attacks}
\end{figure}

Prior work has studied beautification \cite{chen_automatic_2013,alzahrani_deep_2021}, aging effects \cite{kotwal_detection_2020}, and other makeup types such as impersonation and obfuscation \cite{guo_multi-domain_2022}. However, existing datasets often contain limited subjects and exaggerated or unrealistic makeup, reducing their applicability to real-world scenarios. Moreover, some approaches rely on paired-image comparisons with reference identities \cite{rathgeb_detection_2021}, limiting their practicality in open-set settings. Fig.~\ref{fig:makeup_attacks} illustrates examples from public datasets.

In this work, we focus on a more realistic and challenging attack type, referred to as \textit{disguise makeup}. This involves altering facial appearance using both cosmetics and prosthetic materials (e.g., latex, silicone), producing highly natural and often indistinguishable results. Such realism, combined with variations in lighting, pose, image quality, and subject demographics, makes detection particularly challenging.
Despite recent progress in deep learning-based anti-spoofing \cite{sharma_survey_2023}, disguise makeup detection remains underexplored. To address this gap, we introduce a detection framework that does not require paired images and is suitable for open-world scenarios. In addition, we construct a new dataset of live and disguise makeup images collected from public sources, capturing diverse variations in environmental conditions, subject characteristics, and disguise materials.
The proposed approach adopts a two-phase design. First, a global facial model trained with metric learning, enhanced by a whitening transformation, extracts region attention scores using Grad-CAM. These scores then guide a patch-based second phase for localized feature learning and classification, which also incorporates metric learning.

The main contributions are summarized as follows:
\begin{itemize}

\item A generalized disguise makeup presentation attack detection framework is proposed, integrating global facial analysis with region-specific learning to capture fine-grained local patterns and improve robustness across diverse disguise types and conditions. 

\item An attention-guided patch-based learning strategy is introduced, where region attention scores derived from a full-face model guide localized feature learning.

\item Metric learning and style-invariant feature extraction are incorporated to improve embedding separability and robustness to appearance variations under diverse real-world conditions, including disguise styles.

\item A new disguise makeup dataset is constructed, comprising live (genuine) and disguised facial images collected from web and social media sources, covering diverse environmental conditions, subject demographics, and disguise materials to better reflect real-world scenarios.

\end{itemize}
The remainder of this paper is organized as follows. Section 2 reviews related works. Section 3 presents the proposed method. Section 4 reports experimental results and describes the dataset. Section 5 provides discussion, and Section 6 concludes the paper.

\section{Related Works}
Facial makeup poses a significant challenge to face recognition systems, as it can function as a low-cost yet highly effective presentation attack. Early work on makeup detection relied primarily on handcrafted features. Chen et al.~\cite{chen_automatic_2013} extracted shape, texture, and color descriptors from the full face and selected regions, while Liu et al.~\cite{liu_facial_2015} employed entropy measures and gradient orientation pyramids to capture cosmetic-induced texture changes, combined with statistical feature selection and SVM classification. Rasti et al.~\cite{rasti_biologically_2018} proposed a biologically inspired framework based on hierarchical visual processing, combining wavelet-based features with skin tone and gradient descriptors. These approaches, however, showed limited robustness under varying makeup styles, subjects, and acquisition conditions.

To overcome the limitations of handcrafted descriptors, several deep learning–based methods have been introduced. Kotwal et al.~\cite{kotwal_detection_2020} proposed one of the first CNN-based approaches for detecting age-induced makeup attacks by fusing multi-layer features capturing both shape and texture information. Bertacchi et al.~\cite{bertacchi_facial_2019} combined CNNs with the CMYK color model to enhance cosmetic detection by explicitly modeling color information, while Rathgeb et al.~\cite{rathgeb_detection_2021} introduced a differential deep-feature comparison scheme between probe and reference images, trained using synthetically generated makeup attacks. Alzahrani et al.~\cite{alzahrani_deep_2021} investigated deep learning–based makeup detection using transfer learning and semi-supervised approaches to leverage both labelled and unlabelled data.

A major challenge in makeup-based presentation attack detection is generalization to unseen attack types. Although most existing studies do not explicitly focus on disguise makeup, their efforts to address unseen attacks are highly relevant due to the large variability in makeup materials, styles, and application patterns. Several works have therefore reframed PAD as an open-set or anomaly detection problem. Pérez-Cabo et al.~\cite{perez-cabo_deep_2019} proposed a deep metric learning framework that treats presentation attacks as open-set anomalies, while Baweja et al.~\cite{baweja_anomaly_2020} introduced an anomaly detection model trained solely on bona fide samples to improve robustness to unknown attacks. More recently, Huang et al.~\cite{huang_one-class_2024} introduced a one-class anti-spoofing model that simulates spoof cues during training, enabling effective detection of unseen attack types.

Other studies have focused on learning discriminative spoof cues and improving domain generalization. Feng et al.~\cite{feng_learning_2020} proposed a residual learning framework that explicitly models spoof cues at multiple scales, while Guo et al.~\cite{guo_multi-domain_2022} developed a deep learning framework that localizes spoofed regions at the pixel level using multi-branch feature extraction. Chen et al.~\cite{chen_dual-stream_2021} introduced a dual-stream CNN combining standard and central difference convolutions to enhance sensitivity to subtle texture variations, and George et al.~\cite{george_effectiveness_2021} demonstrated the effectiveness of Vision Transformers for cross-database and zero-shot anti-spoofing scenarios. Deb et al.~\cite{deb_look_2021} and Wang et al.~\cite{wang_patchnet_2022} emphasized patch- and region-based feature learning to improve robustness and interpretability, while Zhou et al.~\cite{zhou_instance-aware_2023} and He et al.~\cite{he_center-guided_2025} proposed domain generalization strategies that reduce domain bias through feature normalization and instance-level adaptation. Complementary handcrafted and hybrid approaches have also been explored to enhance texture discrimination. El-Rashidy et al.~\cite{el-rashidy_novel_2025} proposed a texture-based descriptor that captures spatial and directional correlations using multi-block filters, while Radad et al.~\cite{radad_face_2025} introduced a hybrid CNN framework that combines color intensity and handcrafted texture features for robust spoof detection.

Despite these advances, disguise makeup remains insufficiently addressed. Publicly available datasets contain limited makeup-based impersonation samples with restricted diversity in subjects, environments, and materials, hindering realistic evaluation. Moreover, unlike conventional print or replay attacks that introduce global artifacts, disguise makeup produces highly localized and inconsistent facial alterations. This variability complicates detection and underscores the need for methods capable of learning fine-grained, region-specific representations. In this context, patch-based and locally attentive architectures offer a promising direction for robust and interpretable disguise makeup detection.

\section{Proposed Method}
\label{sec:proposed_method}
\begin{figure*}[!t]
    \centering
    \includegraphics[width=\textwidth, height=5in]{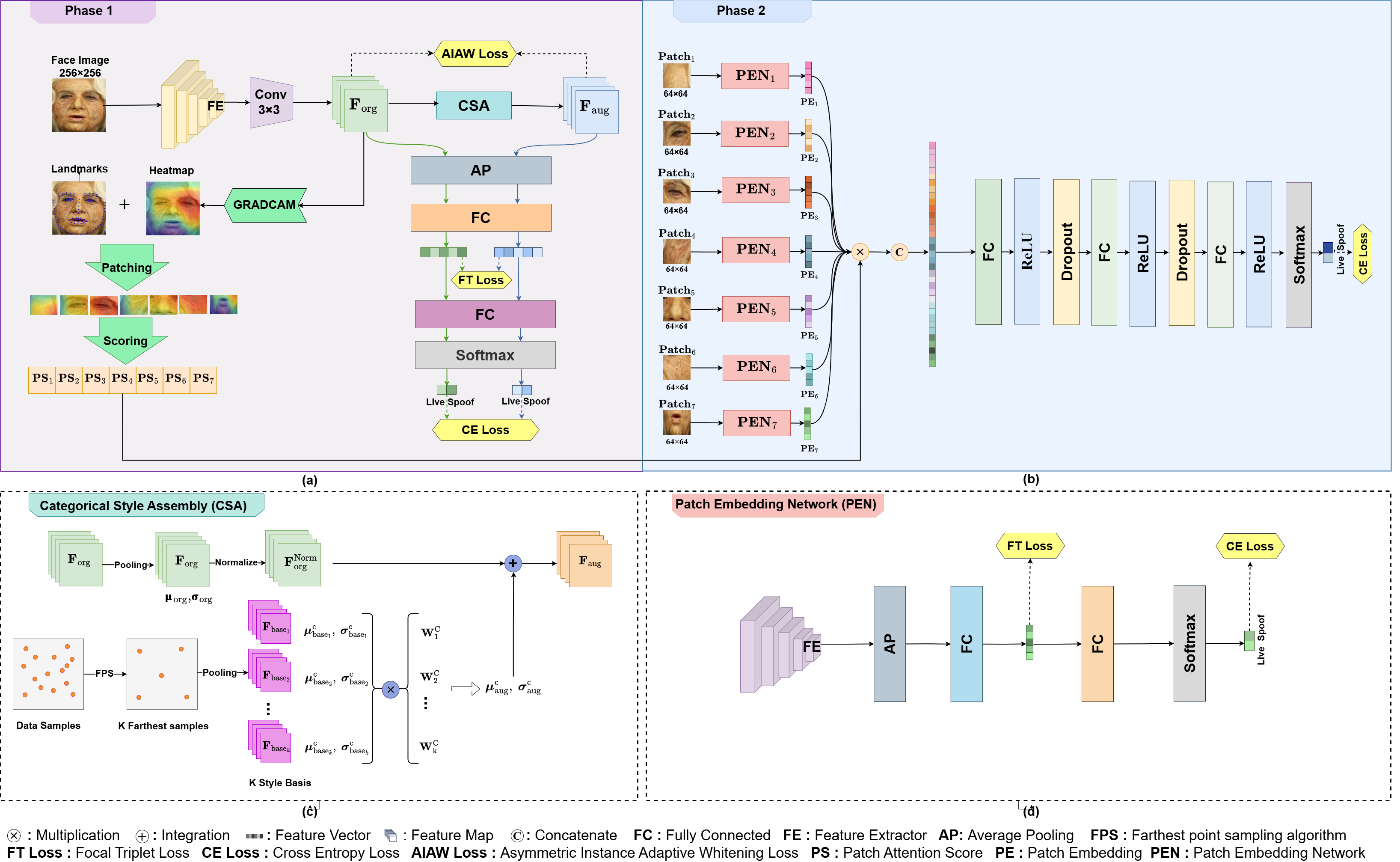} 
    \caption{Overview of the proposed two-phase framework. 
    (a) illustrates Phase 1, where a CNN extracts global features and generates attention scores for predefined facial patches, with robustness enhanced by the CSA module and AIAW-based feature whitening. 
    (b) shows Phase 2, where facial patches are processed independently. Metric learning is applied in both phases to enhance embedding separability. The attention scores computed in Phase 1 are used to weight patch embeddings in Phase 2 for final classification. 
    (c) details the CSA module used in Phase 1, and (d) presents the architecture of the patch embedding network.}
    
    \label{fig:architecture}
\end{figure*}

In this section, we describe the proposed method for generalized disguise makeup attack detection. As shown in Fig.~\ref{fig:architecture}, the approach consists of two main phases, preceded by a preprocessing step. In the first phase, the entire facial image is analyzed to estimate attention scores for predefined facial patches. In the second phase, a patch-based strategy is employed to extract localized features. The attention scores derived from the first phase are then used to weight these patches, enabling the construction of a discriminative embedding for final prediction.
\subsection{Preprocessing}
To prepare the data for the proposed architecture, the following steps were performed: (1) face detection and alignment, and (2) facial landmark detection for extracting relevant patches.
\subsubsection{Face Detection and Alignment}
\begin{figure}[!]
    \centering
    \includegraphics[
        width=0.55\columnwidth,
        height=2cm,
        keepaspectratio
    ]{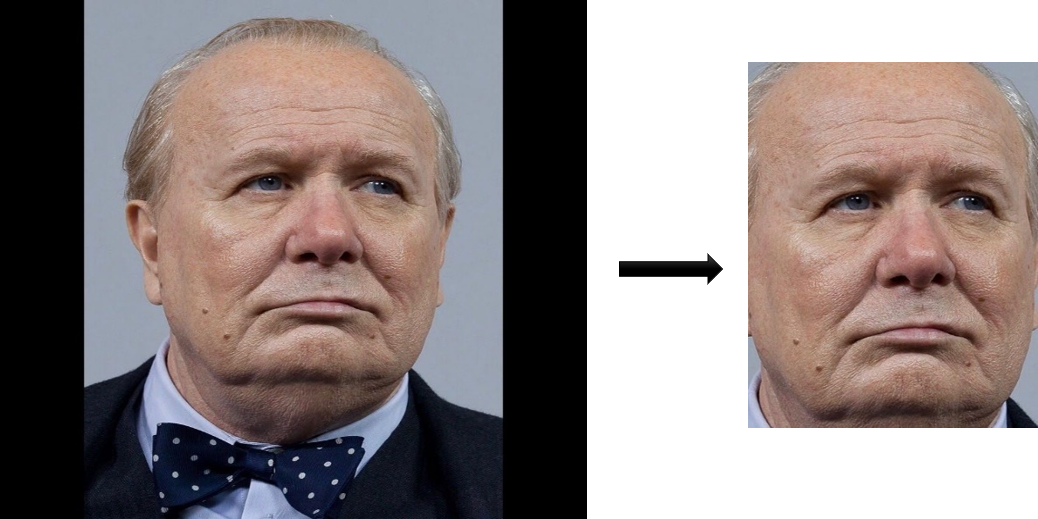} 
    \caption{Example of an image in the dataset (left) and its preprocessed version (right).}
    \label{fig:preprocess}
\end{figure}

Face detection was applied to generate inputs for the full-face network (Phase 1) and to extract patches for Phase 2. Alignment was performed to normalize facial pose and reduce bias toward orientation, which is common in face presentation attack detection.
We used DeepFace\footnote{https://github.com/serengil/deepface}, a lightweight Python framework providing multiple pre-trained face detectors. MTCNN was used as the primary detector, and YOLOv8 was applied as a fallback when detection failed. Face alignment was also performed using this framework. Fig.~\ref{fig:preprocess} shows an example before and after preprocessing.
\subsubsection{Landmark Detection and Patch Extraction}
\begin{figure}[!]
    \centering
    \includegraphics[height=1in,keepaspectratio]{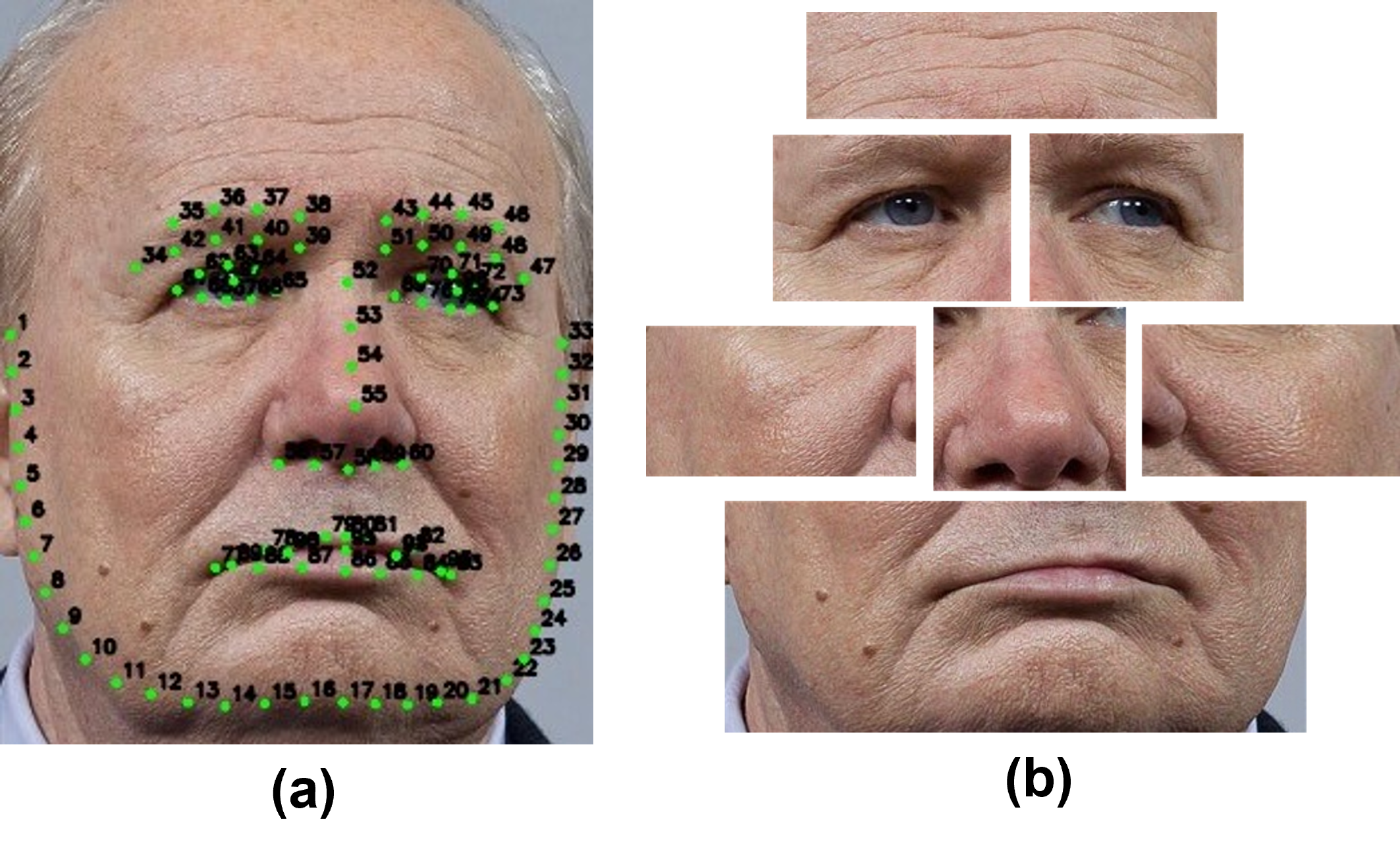} 
    \caption{Facial landmarks detected using SPIGA (a) and the corresponding patches extracted based on these landmarks (b). }
    \label{fig:spiga}
\end{figure}

Facial landmarks are extracted using the SPIGA framework \cite{prados-torreblanca_shape_2022}, applied to original-resolution face crops from the previous step to avoid resolution loss. It provides robust localization under challenging conditions such as heavy makeup, occlusions, and pose variations. Based on the detected landmarks, seven facial regions—forehead, left eye, right eye, left cheek, nose, right cheek, and mouth–chin—are extracted and used as input patches for the second-phase subnetworks. Fig.~\ref{fig:spiga} illustrates the results, where (a) shows detected landmarks and (b) the corresponding extracted patches.

\subsection{Phase 1: Full-Face-Based Attention Score Extraction Network}
\subsubsection{Full-face Based Network}
The full-face network, as shown in Fig.~\ref{fig:architecture}(a), learns facial representations that are later used for attention score extraction. A pre-trained CNN is fine-tuned to extract feature maps from the input face image.

Input images are resized to a resolution of $H \times W$ (set to $256 \times 256$ in our experiments) and passed through the feature extractor, followed by a convolutional layer that reduces the channels to 640, producing $F_{\text{org}}$. To improve robustness to style variations caused by diverse capture conditions, a style augmentation module (CSA; Fig.~\ref{fig:architecture}(c)) \cite{zhou_instance-aware_2023} generates augmented features $F_{\text{aug}}$, reducing style bias and enhancing generalization.
 
To reduce the influence of domain-specific styles and improve discrimination between live and disguise makeup images, we employ the Adaptive Asymmetric Instance-Aware Whitening (AIAW) module \cite{zhou_instance-aware_2023}. This module operates on both the original and style-augmented feature maps ($F_{\text{org}}$ and $F_{\text{aug}}$) and suppresses style-sensitive components by enforcing decorrelation in the feature space.

The AIAW loss is defined as:
\begin{equation}
L_{\text{AIAW}} = \sum_{k_c \in \{k_l, k_m\}} 
\sum_{t \in \{\text{org}, \text{aug}\}} 
\mathbb{E} \Big[ \big\| \Sigma_t \odot M(k_c) \big\| \Big]
\end{equation}
where $\Sigma_t$ denotes the covariance matrix of normalized features, $M(k_c)$ is a binary mask selecting style-sensitive components for each class, with $k_c$ controlling the proportion of selected elements for live ($k_l$) and makeup ($k_m$) samples, and $\odot$ represents element-wise multiplication. The loss suppresses domain-specific variations while preserving discriminative facial features.

Subsequently, the features ($F_{org}$ and $F_{aug}$) are passed through an average pooling layer and a fully-connected layer to produce a 64-dimensional embedding vector for each of them. A focal triplet loss is applied to these embeddings to encourage embeddings of live faces to be closer to each other than to makeup faces. Triplets are generated online using a random hard negative mining strategy, where negatives are randomly sampled from margin-violating examples:
\begin{equation}
L_{TF}(\theta) = \sum_{i=1}^b
\max \Bigg( 0,\ e^{\frac{D(a_i,p_i)}{\sigma}} - e^{\frac{D(a_i,n_i)}{\sigma}} + m \Bigg)
\label{eq:ltf}
\end{equation}
where $a_i$, $p_i$, and $n_i$ denote anchor, positive, and negative samples, $D(\cdot,\cdot)$ represents the squared Euclidean distance, $m$ is the margin, and $\sigma$ is a scaling factor.

Finally, for supervised classification, the embeddings are fed into a fully connected layer with two neurons, and cross-entropy loss $L_{CE}$ is applied. The total loss is formulated as:
\begin{equation}
\begin{split}
L_{\text{Total}} &= \alpha_1 L_{\text{AIAW}}^{\text{org}}
                 + \beta_1 L_{\text{TF}}^{\text{org}}
                 + \gamma_1 L_{\text{CE}}^{\text{org}} \\
                 &\quad + \alpha_2 L_{\text{AIAW}}^{\text{aug}}
                 + \beta_2 L_{\text{TF}}^{\text{aug}}
                 + \gamma_2 L_{\text{CE}}^{\text{aug}}
\end{split}
\label{eq:total_loss}
\end{equation}

where $\alpha_i$, $\beta_i$, and $\gamma_i$ control the contribution of each component in the final loss.
\subsubsection{Attention Scores Extraction Process}
After training the full-face network, Grad-CAM is applied to highlight the important regions of an image for a given prediction by leveraging the gradients of the last convolutional layer and projecting them onto the corresponding activation maps. So, the gradients of the final convolutional layer of the full-face model are extracted during the backward pass. These gradients are then combined with the activation maps to generate a heatmap representing the most influential regions of the image. To retain only the positively contributing areas for the model's prediction, a ReLU is applied to the heatmap, which is then normalized. Finally, the resulting heatmap is resized to match the input resolution ($H \times W$).

To compute the attention score for each facial patch, the corresponding region for each patch must first be identified. Using the facial landmarks extracted from the original image (before resizing to $H \times W$), the scaling factors for width and height are calculated as follows:

\begin{align}
\text{scale\_width} &= \frac{W}{\text{main\_width}} \\[8pt]
\text{scale\_height} &= \frac{H}{\text{main\_height}}
\end{align}

The new landmark coordinates corresponding to the resized image are then obtained using:

\begin{equation}
\begin{cases}
\text{new\_landmark}_x^i = \text{old\_landmark}_x^i \times \text{scale\_width} \\[8pt]
\text{new\_landmark}_y^i = \text{old\_landmark}_y^i \times \text{scale\_height}
\end{cases}
\end{equation}

Based on these updated landmarks, the region corresponding to each patch is identified on the heatmap, resulting in a patch-specific heatmap. For each patch, the top $k\%$ of heatmap values are selected and summed to compute the attention score, which reflects the contribution of the most influential regions. This score is then normalized by dividing by the number of pixels in the top $k\%$, ensuring independence from patch size while emphasizing the relative importance of each patch. The value of $k$ is determined empirically through ablation studies, which are presented in the following section.

\subsection{Phase 2: Attention-Guided Patch-Based Network}

In this phase, a set of patch-specific subnetworks is employed to capture fine-grained and localized facial details. For each facial patch, an independent CNN-based network with common structure is utilized to extract region-specific features, as illustrated in Fig.~\ref{fig:architecture}(b). This design enables the model to focus on localized cues rather than relying solely on global patterns, thereby reducing the risk of overfitting to holistic facial representations. 

This approach is particularly beneficial for \textit{disguise makeup detection}, where alterations are not uniform across facial regions. In different instances, some patches may be heavily modified while others remain less affected, requiring patch-wise discriminative feature learning.

Each patch is resized to $h \times w$ (i.e., $64 \times 64$) and processed by its corresponding feature extractor. The extracted features are then projected into an embedding vector of dimension $16$ via a fully connected layer. To optimize these embeddings, we adopt a metric-learning approach using a triplet focal loss with online random hard negative mining, as described in Eq.~\eqref{eq:ltf}. This loss encourages embeddings from live images to form a compact cluster while pushing apart embeddings of makeup instances. In other words, the objective is not only to learn a decision boundary but also to structure the embedding space effectively, improving generalization. This property is especially important for disguise makeup detection, where localized alterations introduce large intra-class variations.

In addition to $\mathcal{L}_{TF}$, a cross-entropy loss, $\mathcal{L}_{CE}$, is applied for patch-level classification, providing explicit supervision at the class level. The overall loss for patch $i$ is defined as:
\begin{equation}
\mathcal{L}_{\text{Patch}_i} = \alpha \mathcal{L}_{CE} + \beta \mathcal{L}_{TF},
\end{equation}
where $\alpha$ and $\beta$ control the contribution of each term.

Finally, to integrate global attention into the patch-based framework, the attention scores computed in Phase~1 are applied to the embeddings. Each embedding is weighted by its corresponding attention score, emphasizing the most informative patches. The weighted embeddings are concatenated and passed through a Multi-Layer Perceptron (MLP) for the final classification. This final network is optimized using the cross-entropy loss.
\section{Experimental Results}

\subsection{Datasets}
Due to the lack of a dedicated dataset for disguise makeup detection, we construct a new dataset as the primary benchmark. For additional evaluation, we also use the SiW-Mv2 dataset \cite{guo_multi-domain_2022}, which includes makeup-based attacks partially aligned with this task. Moreover, existing datasets contain few concealment or impersonation samples created with conventional cosmetics, failing to capture the complexity of disguise makeup.

\subsubsection{Collected Disguise Makeup Dataset}
Existing face presentation attack datasets often lack subject diversity and contain limited disguise samples, restricting their applicability in real-world scenarios \cite{costa-pazo_challenges_2019}.

\begin{figure}[!t]
    \centering
    \includegraphics[width=0.75\columnwidth,height=1.75in]{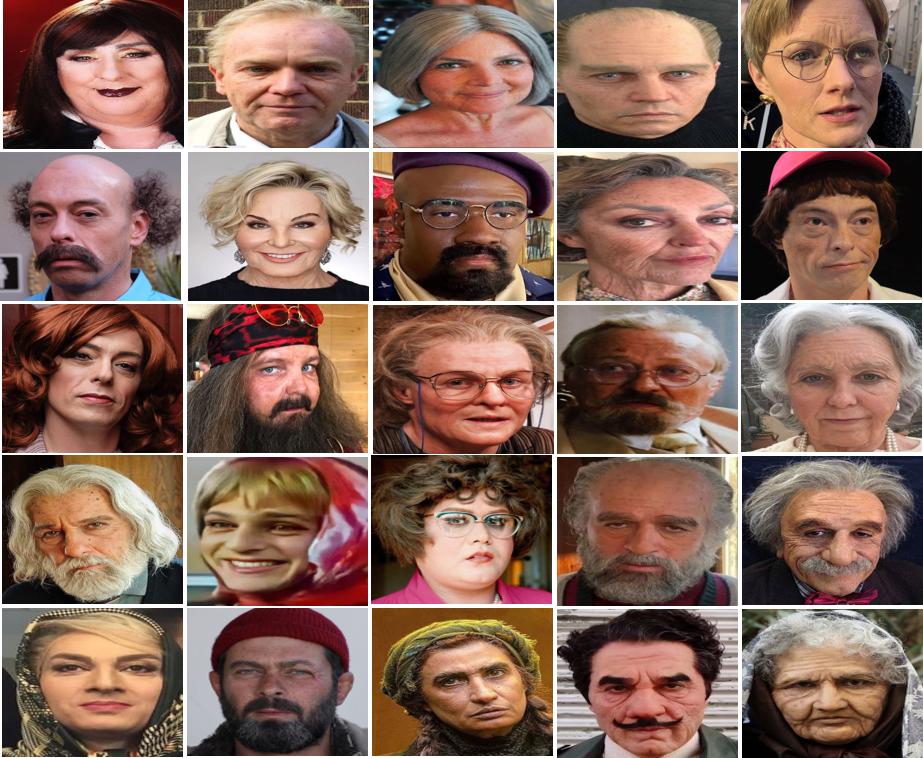} 
    \caption{Examples of makeup presentation attack samples from collected disguise makeup dataset. }
    \label{fig:grim_makeup_attacks}
\end{figure}

To address this, we constructed a dataset of disguise makeup and live facial images from public sources. Disguise makeup images were collected from publicly available web images and public Instagram pages of professional makeup artists. The dataset includes 635 disguise subjects (853 images), all with frontal views. Among them, 303 subjects are Iranian and 332 are from other nationalities, ensuring demographic diversity. Only realistic human disguises were retained. Non-human or fantasy transformations were excluded.
The dataset covers a wide range of disguise techniques, including prosthetic-based and conventional cosmetic approaches, using materials such as latex and silicone, with variations in illumination, background, camera conditions, distance, image quality, and demographics. These factors increase task difficulty and realism. Examples are shown in Fig.~\ref{fig:grim_makeup_attacks}.

For the live class, 719 images of different subjects were collected from public sources, including Instagram, web images, Flickr\footnote{https://www.kaggle.com/datasets/arnaud58/flickrfaceshq-dataset-ffhq}, BookClub \cite{selitskiy_bookclub_2021}, and Celeb-A \cite{liu_deep_2015}. The resulting dataset provides a challenging benchmark for disguise makeup detection.

A summary is given in Table~\ref{tab:dataset_summary}.

\begin{table}[ht]
\centering
\small
\setlength{\tabcolsep}{4pt}
\renewcommand{\arraystretch}{1.3}
\caption{Summary of the collected disguise makeup dataset.}
\begin{tabular}{p{2.8cm} p{5.5cm}}
\hline
\textbf{Property} & \textbf{Description} \\
\hline
Disguise subjects & 635 \\
Live subjects & 719 \\
View constraint & Frontal faces only \\
Variations & Illumination, background, camera, distance, quality, disguise styles and application techniques \\
Demographic diversity & Age, gender, ethnicity \\
Disguise sources & Web and Instagram \\
Live sources & Instagram, Web, Flickr, BookClub,Celeb-A \\
Image resolution & 
Live: $40 \le W \le 838$, $55 \le H \le 1111$; 
Disguise: $34 \le W \le 1111$, $48 \le H \le 1430$ \\
\hline
\end{tabular}
\label{tab:dataset_summary}
\end{table}

 \begin{figure}[!t]
    \centering
    \includegraphics[width=0.75\columnwidth,height=1.75in]{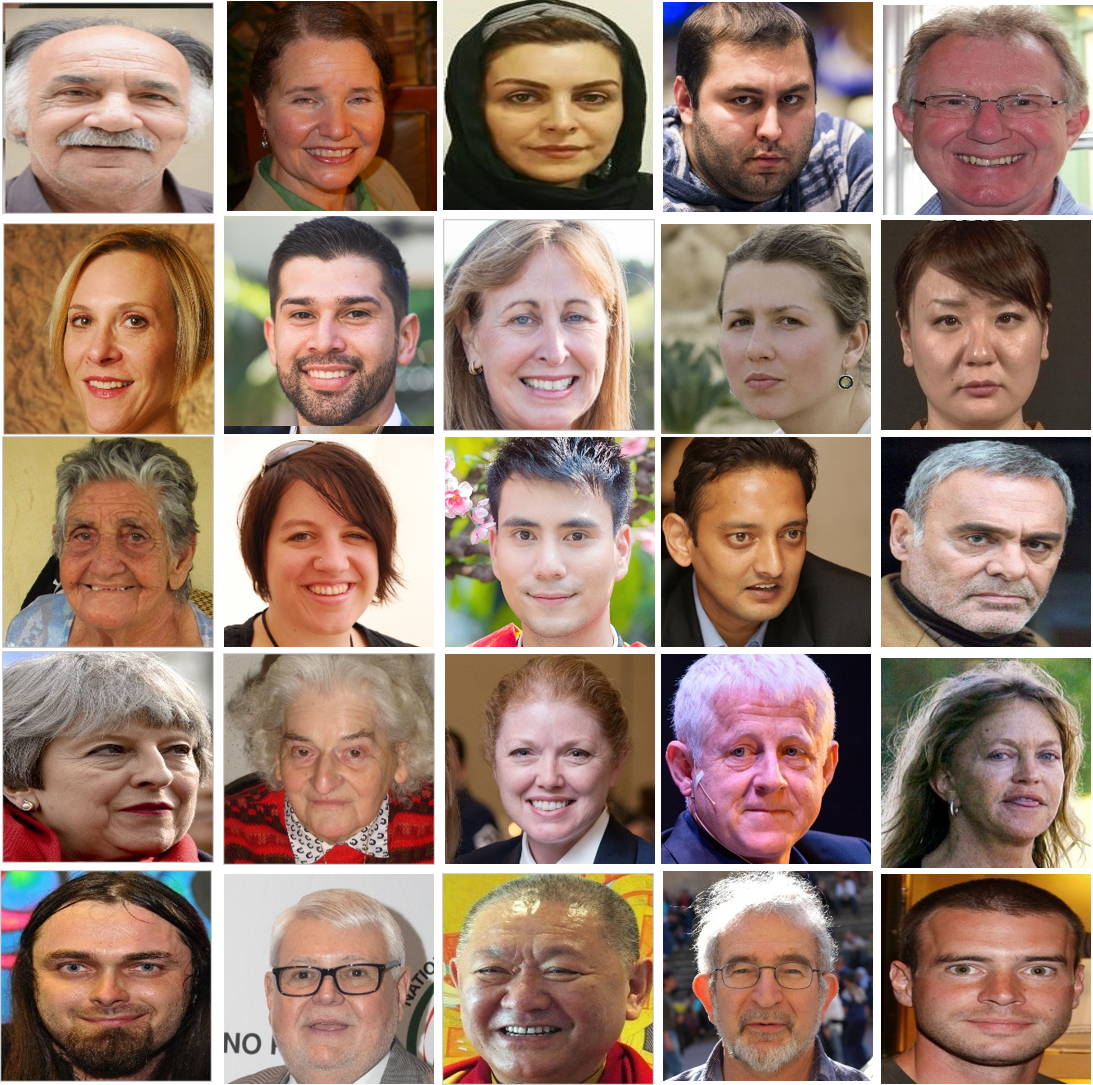} 
    \caption{Examples of live samples from collected disguise makeup dataset.}
    \label{fig:no-grim_makeup_attacks}
\end{figure}
 
\begin{figure*}[!t]
    \centering
    \begin{subfigure}[b]{0.32\textwidth}
        \centering
        \includegraphics[width=\textwidth,height=1.1in]{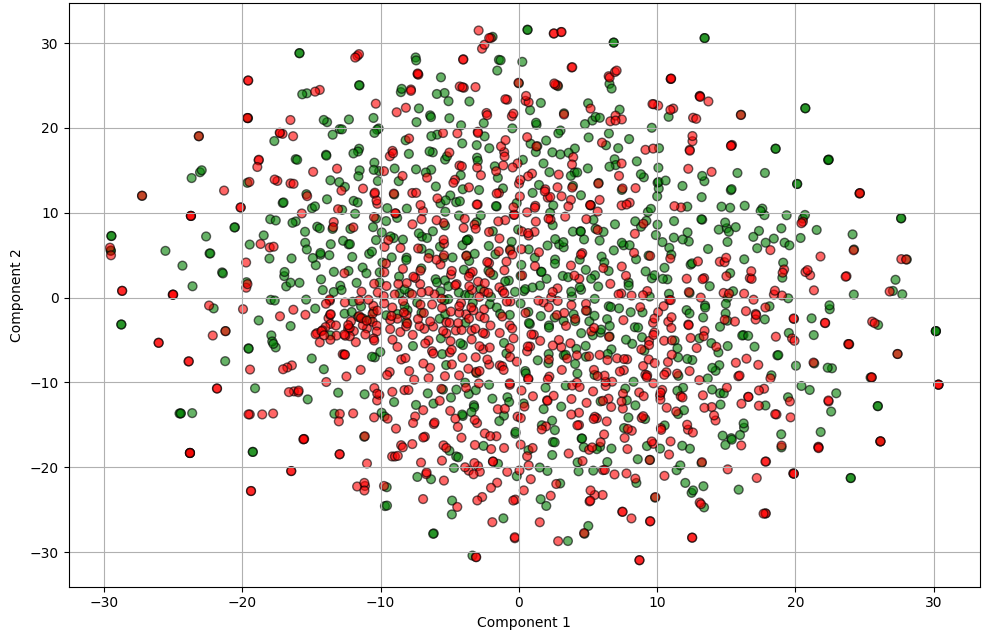}
        \caption{}
        \label{fig:sub1}
    \end{subfigure}
    \hfill
    \begin{subfigure}[b]{0.32\textwidth}
        \centering
        \includegraphics[width=\textwidth,height=1.1in]{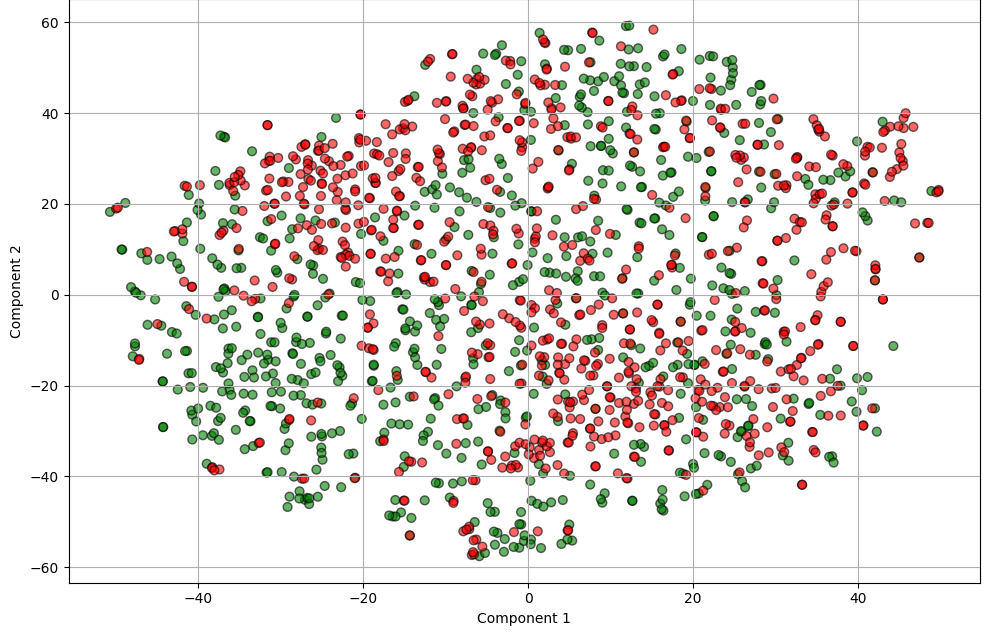}
        \caption{}
        \label{fig:sub2}
    \end{subfigure}
    \hfill
    \begin{subfigure}[b]{0.32\textwidth}
        \centering
        \includegraphics[width=\textwidth,height=1.1in]{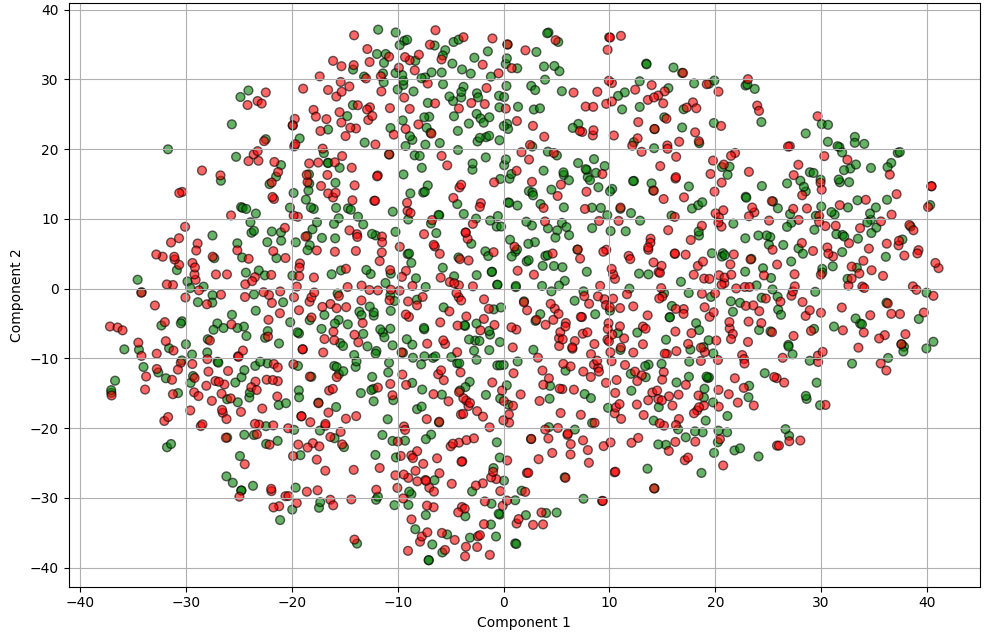}
        \caption{}
        \label{fig:sub3}
    \end{subfigure}
    
\caption{t-SNE visualization of feature embeddings extracted using (a) LightCNN, (b) FaceNet, and (c) ArcFace, where the points correspond to samples from the collected dataset. Green points represent live faces, while red points denote disguise makeup samples. The disguise samples exhibit high intra-class variability without forming compact clusters, and their overlap with live samples indicates low inter-class separability. These results demonstrate the challenging nature of the dataset and highlight the limitations of conventional face recognition embeddings for disguise makeup detection.}
    \label{fig:all_images}
\end{figure*}

To analyze the complexity of the collected dataset, we visualize embeddings from existing face recognition models. Fig.~\ref{fig:all_images} shows t-SNE projections of embeddings extracted from LightCNN, FaceNet, and ArcFace.

\subsubsection{Spoof-in-the-Wild With Multiple Attacks (SiW-Mv2) Dataset}

The SiW-Mv2 dataset includes 14 spoofing categories, ranging from print and replay attacks to various mask types, impersonation makeup, and occlusions. It contains 785 live videos from 493 subjects and 915 spoof videos from 600 subjects, offering high diversity in attack patterns.

Among these, three categories—cosmetics, impersonation, and obfuscation—are makeup-based and thus most relevant to this work, particularly impersonation. Since disguise makeup represents a specific type of spoofing, we adopt Protocol 1 (known-attack scenario) for evaluation.
\subsection{Implementation Details}
The proposed method is implemented in PyTorch. The framework utilizes MobileNetV2~\cite{sandler_mobilenetv2_2018} as the feature extractor for both the full-face-based network and the patch-specific subnetworks, which are trained using the Adam optimizer with an initial learning rate of 0.001. In the full-face-based network, the number of base styles is set to 64, and the feature maps extracted from the convolutional layer following the backbone have a dimension of 640. Additionally, the selection thresholds for the binary selection mask $M$ in the first phase are set to $k_l = 0.09\%$ and $k_m = 0.06\%$. All experiments were conducted on an NVIDIA T4 GPU. The remaining parameter settings are summarized in Table~\ref{tab:training_parameters}.
\begin{table}[t]
\centering
\caption{Training settings and loss parameters for the full-face and patch-based networks.}
\label{tab:training_parameters}
\begin{adjustbox}{max width=\columnwidth}
\large   
\begin{tabular}{>{\centering\arraybackslash}m{3.2cm} 
                >{\centering\arraybackslash}m{1.5cm} 
                >{\centering\arraybackslash}m{2cm} 
                >{\centering\arraybackslash}m{6cm}}
\hline
\textbf{Network} & \textbf{Epochs} & \textbf{Input Size} & \textbf{Loss Parameters} \\
\hline
\makecell{Full-face \\ network} & 30 & $256 \times 256$ & 
\makecell{$\sigma = 2, \; m = 0.6,$ \\ $\alpha_1=1, \alpha_2=1,$ \\ $\beta_1=0.1, \beta_2=0.1,$ \\ $\gamma_1=1, \gamma_2=1$} \\
\makecell{Patch-based \\ networks} & 20 & $64 \times 64$ & 
\makecell{$\sigma = 1.5, \; m = 1,$ \\ $\alpha=1,$ \\ $\beta=0.1$} \\
\hline
\end{tabular}
\end{adjustbox}
\end{table}

\subsection{Ablation Studies}
In this section, we analyze the contribution of different components and design choices of the proposed framework through ablation experiments. To this end, 80\% of the subjects in the collected dataset were randomly assigned to the training set, while the remaining 20\% were used for evaluation. The different architectures were then trained and evaluated based on this split, and the best-performing configuration was selected as the proposed method.
\subsubsection{Evaluation of Facial Landmark Detection Models}
Accurate landmark detection is essential for reliable patch extraction in the proposed framework. To assess the robustness of facial landmark detection on disguise makeup images, we compare two facial landmark detectors: dlib \cite{king_dlib-ml_2009} and SPIGA \cite{prados-torreblanca_shape_2022}.

dlib uses a 68-point landmark detector based on an ensemble of regression trees, while SPIGA employs a graph attention–based architecture with 98 landmark points.

Table~\ref{tab:landmark_performance} shows landmark detection success rates on the collected dataset. dlib fails on several images, particularly under heavy disguise makeup, and also misses a few live samples. In contrast, SPIGA achieves 100\% detection success on both live and disguise images, demonstrating superior robustness under challenging conditions.

\begin{table}[ht] \centering \caption{Performance comparison of dlib and SPIGA on the collected dataset.} \renewcommand{\arraystretch}{1.5} 
\begin{tabular}{lcc|cc} \hline \multirow{2}{*}{Method} & \multicolumn{2}{c|}{Makeup} & \multicolumn{2}{c}{Live} \\ & Success & Failure & Success & Failure \\ \hline dlib & 775 & 78 & 712 & 7 \\ SPIGA & 853 & 0 & 719 & 0 \\ \hline \end{tabular} \label{tab:landmark_performance} \end{table}
Based on these results, SPIGA was selected as the landmark detection module for all subsequent experiments.

\subsubsection{Full-Face Based Models}
Since the proposed method leverages both full-face and patch-based features, we first evaluate models based solely on full-face features. Four variants are considered, and results are reported in Table~\ref{tab:fullface_ablation}. All models use a pre-trained MobileNetV2 as the feature extractor.

Results show that adding either the triplet focal loss or the CSA and AIAW modules to the Cross-Entropy baseline model improves performance, while combining all components yields the best results. Specifically, CSA expands the feature space by generating new styles, improving robustness to appearance variations. AIAW suppresses domain-specific style information and enhances discriminative feature learning for distinguishing live and disguised faces. The triplet focal loss encourages compact embeddings for live samples while separating disguised samples, improving representation quality. All models are trained for 30 epochs.

To select the optimal feature extractor for the first phase, we further compare MobileNetV2 with InceptionResNetV1 in Table~\ref{tab:feature_extractor_comparison}. InceptionResNetV1, a widely used face recognition backbone, is evaluated both without fine-tuning (pre-trained only) and with fine-tuning during training. In both cases, MobileNetV2 achieves better performance.

This can be attributed to the difference between face recognition and disguise detection tasks, where face recognition networks are designed to cluster identities, whereas disguise makeup alters identity-related cues. In contrast, MobileNetV2 \cite{sandler_mobilenetv2_2018} provides a more suitable balance between efficiency and feature representation for this task.

\begin{table}[t]
\centering
\renewcommand{\arraystretch}{1.2}
\caption{Ablation results for full-face models.}
\label{tab:fullface_ablation}
\begin{adjustbox}{max width=\columnwidth}
\begin{tabular}{>{\centering\arraybackslash}m{3.6cm} 
                >{\centering\arraybackslash}m{1.5cm} 
                >{\centering\arraybackslash}m{1.5cm} 
                >{\centering\arraybackslash}m{1.5cm} 
                >{\centering\arraybackslash}m{1.5cm}}
\hline
\textbf{Model} & 
\makecell{\textbf{Accuracy} \\ (\%)} & 
\makecell{\textbf{BPCER} \\ (\%)} & 
\makecell{\textbf{APCER} \\ (\%)} & 
\makecell{\textbf{ACER} \\ (\%)} \\
\hline
\makecell{Cross-Entropy} & 86.90 & 16.43 & 9.30 & 12.87 \\
\makecell{Cross-Entropy  + CSA + AIAW} & 87.63 & 15.75 & \textbf{8.52} & 12.14 \\
\makecell{Cross-Entropy  + Triplet Focal Loss} & 88.00 & 13.69 & 10.07 & 11.88 \\
\makecell{Cross-Entropy  + CSA + AIAW \\ + Triplet Focal Loss} & \textbf{89.45} & \textbf{12.32} & \textbf{8.52} & \textbf{10.42} \\
\hline
\end{tabular}
\end{adjustbox}
\end{table}

\renewcommand{\arraystretch}{1.3} 

\begin{table}[t]
\centering
\caption{Comparison of different feature extractors for the full-face model.}
\label{tab:feature_extractor_comparison}
\begin{adjustbox}{max width=\columnwidth}
\large
\begin{tabular}{>{\centering\arraybackslash}m{3.5cm} 
                >{\centering\arraybackslash}m{1.5cm} 
                >{\centering\arraybackslash}m{1.8cm} 
                >{\centering\arraybackslash}m{1.8cm} 
                >{\centering\arraybackslash}m{1.8cm} 
                >{\centering\arraybackslash}m{1.8cm}}
\hline
\makecell{\textbf{Feature Extractor}} & 
\makecell{\textbf{Fine-} \\ \textbf{tuned}} & 
\makecell{\textbf{Accuracy} \\ (\%)} & 
\makecell{\textbf{BPCER} \\ (\%)} & 
\makecell{\textbf{APCER} \\ (\%)} & 
\makecell{\textbf{ACER} \\ (\%)} \\
\hline
\makecell{InceptionResNetV1} & -- & 80.00 & 16.43 & 24.03 & 20.23 \\
\makecell{InceptionResNetV1} & \checkmark & 82.54 & 18.49 & 16.27 & 17.38 \\
\makecell{MobileNetV2} & \checkmark & \textbf{89.45} & \textbf{12.32} & \textbf{8.52} & \textbf{10.42} \\
\hline
\end{tabular}
\end{adjustbox}
\end{table}

\subsubsection{Patch-Based Networks}

In this section, the patch-based model is analyzed and evaluated. As previously described, each facial patch is processed by an independent feature extractor. The base architecture comprises seven MobileNetV2 subnetworks, each trained using the cross-entropy loss. The resulting patch embeddings are concatenated and passed through a multi-layer perceptron (MLP) for final classification.

Table~\ref{tab:patch_ablation_loss} presents the impact of incorporating the triplet focal loss in addition to cross-entropy for training the patch networks. The results show that incorporating metric learning improves the overall results. Furthermore, ResNet18 was also evaluated as a feature extractor. However, MobileNetV2 consistently yielded better performance and was used for all subsequent experiments. All patch-based models were trained for 20 epochs.

To validate the effectiveness of concatenating patch embeddings for final decision-making, a majority-voting scheme was also implemented. In this method, the predicted label for each of the seven patches was obtained, and the label with the most votes was selected as the final decision. Table~\ref{tab:patch_fusion_comparison} shows that concatenation followed by the MLP yields superior performance compared to majority voting, likely due to the MLP’s ability to account for noise in individual patch predictions.

Finally, the impact of attention mechanisms on patch embeddings is investigated. As shown in Table~\ref{tab:patch_attention_comparison}, the first row presents the best-performing patch-based network identified in the ablation study (Table~\ref{tab:patch_fusion_comparison}). Building on this baseline, three attention mechanisms are evaluated: a four-head attention layer, a pre-trained DeiT transformer, and the proposed Grad-CAM-based attention mechanism (with $k=50\%$, as determined in a subsequent ablation study). The results demonstrate that the proposed method achieves the best performance, significantly outperforming the alternative approaches.

\begin{table}[t]
\centering
\caption{Performance Comparison of Patch-Based Models with Different Loss Configurations and Feature Extractors
}
\label{tab:patch_ablation_loss}
\begin{adjustbox}{max width=\columnwidth} 
\large 
\renewcommand{\arraystretch}{1.6} 
\begin{tabular}{l c c c c}
\hline
\makecell[c]{\textbf{Patch Model}} & 
\makecell[c]{\textbf{Accuracy} \\ (\%)} & 
\makecell[c]{\textbf{BPCER} \\ (\%)} & 
\makecell[c]{\textbf{APCER} \\ (\%)} & 
\makecell[c]{\textbf{ACER} \\ (\%)} \\
\hline
MobileNetV2 + Cross-Entropy & 89.75 & 13.69 & 7.52 & 10.61 \\
MobileNetV2 + Cross-Entropy + Triplet Focal Loss & \textbf{90.36} & \textbf{13.01} & \textbf{6.98} & \textbf{10.00} \\
ResNet18 + Cross-Entropy + Triplet Focal Loss & 88.85 & 15.06 & 8.06 & 11.56 \\
\hline
\end{tabular}
\end{adjustbox}
\end{table}

\begin{table}[t]
\centering
\renewcommand{\arraystretch}{1.4}
\caption{Comparison of patch fusion methods.}
\label{tab:patch_fusion_comparison}
\resizebox{\columnwidth}{!}{
\begin{tabular}{l c c c c}
\hline

\textbf{Fusion Method} &
\makecell{\textbf{Accuracy} \\ (\%)} &
\makecell{\textbf{BPCER} \\ (\%)} &
\makecell{\textbf{APCER} \\ (\%)} &
\makecell{\textbf{ACER} \\ (\%)}\\
\hline
Majority Voting & 89.45 & 15.06 & 6.98 & 11.02 \\
Concatenated Embeddings + MLP & \textbf{90.36} & \textbf{13.01} & \textbf{6.98} & \textbf{10.00} \\
\hline
\end{tabular}
}
\end{table}
\begin{table}[t]
\centering
\caption{Comparison of attention mechanisms on patch embeddings.}
\label{tab:patch_attention_comparison}
\resizebox{\columnwidth}{!}{
\begin{tabular}{l c c c c}
\hline

\textbf{Attention Method} &
\makecell{\textbf{Accuracy} \\ (\%)} &
\makecell{\textbf{BPCER} \\ (\%)} &
\makecell{\textbf{APCER} \\ (\%)} &
\makecell{\textbf{ACER} \\ (\%)}\\
\hline
No Attention & 90.36 & 13.01 & 6.98 & 10.00 \\
Four-Head Attention & 90.2 & 14.0 & 5.74 & 10.1 \\
Pre-trained DeiT & 90.9 & 13.10 & 5.74 & 9.4 \\
Proposed Method & \textbf{92.46} & \textbf{10.27} & \textbf{5.37} & \textbf{7.82} \\
\hline
\end{tabular}
}
\end{table}

\subsubsection{Optimal $k\%$ for Grad-CAM-Based Patch Attention Scores}\label{subsubsec:gradcam_k}

In this section, Grad-CAM is applied to the full-face network to generate heatmaps for the input images. These heatmaps are then used to compute patch-level attention scores. Experiments were conducted to determine the optimal percentage $k\%$ of top heatmap values to consider for each patch, and the results are reported in Table~\ref{tab:gradcam_k}. As shown, using the top 50\% of values provides the best performance; hence, $k=50\%$ is selected for the final model.
\renewcommand{\arraystretch}{1.3}
\begin{table}[t]
\centering
\caption{Evaluation of different $k\%$ values for Grad-CAM-based patch attention scores.}
\label{tab:gradcam_k}
\resizebox{\columnwidth}{!}{%
\begin{tabular}{c c c c c}
\hline

\textbf{$k\%$} & \textbf{Accuracy (\%)} & \textbf{BPCER (\%)} & \textbf{APCER (\%)} & \textbf{ACER (\%)} \\
\hline
40 & 91.26 & 8.21 & 9.13 & 8.67 \\
50 & \textbf{92.46} & \textbf{10.27} & \textbf{5.37} & \textbf{7.82} \\
60 & 91.56 & 11.64 & 5.91 & 8.77 \\
\hline
\end{tabular}%
}
\end{table}

\subsection{Results and Comparisons}
In this section, we first present the results of 5-fold cross-validation with subject-disjoint folds on the collected disguise makeup dataset using the proposed architecture. Subsequently, we evaluate the performance of the proposed architecture and previous methods on attack scenarios from the SiW-Mv2 dataset. Finally, a comparison with prior works on the collected dataset is reported.
Table ~\ref{tab:5fold_results} shows the results of subject-disjoint 5-fold cross-validation of the final architecture on the collected disguise makeup dataset. As observed, the proposed architecture demonstrates consistent performance across all folds.

\begin{table}[t]
\centering
\caption{Subject-disjoint 5-fold cross-validation results on the collected disguise makeup dataset.}
\label{tab:5fold_results}
\begin{tabularx}{\columnwidth}{>{\centering\arraybackslash}m{1.2cm} 
                             >{\centering\arraybackslash}X 
                             >{\centering\arraybackslash}X 
                             >{\centering\arraybackslash}X 
                             >{\centering\arraybackslash}X 
                             >{\centering\arraybackslash}X }
\hline
\textbf{Fold} & \textbf{Accuracy (\%)} & \textbf{BPCER (\%)} & \textbf{APCER (\%)} & \textbf{ACER (\%)} & \textbf{EER (\%)} \\
\hline
1 & 90.06 & 12.36 & 6.84 & 9.60 & 11.2 \\
2 & 92.35 & 10.95 & 4.51 & 7.73 & 8.90 \\
3 & 89.06 & 10.95 & 10.90 & 10.93 & 10.9 \\
4 & 91.48 & 11.64 & 6.01 & 8.82 & 9.5 \\
5 & 92.35 & 9.58 & 5.95 & 7.77 & 8.3 \\
\hline
\textbf{Mean ± Std} & 91.06 ± 1.30 & 11.10 ± 0.92 & 6.84 ± 2.16 & 8.97 ± 1.20 & 9.76 ± 1.12 \\
\hline
\end{tabularx}
\end{table}
A comparison between the proposed architecture and prior work on face presentation attack detection was conducted using the dataset collected in this study. Table~\ref{tab:collected_dataset_comparison} reports the performance of the proposed method in comparison with the ViTranZFAS model \cite{george_effectiveness_2021}. As observed, the proposed architecture consistently outperforms the baseline across all reported metrics. Notably, the EER is reduced from $15.20\%$ for ViTranZFAS to $9.76\%$ for the proposed method, highlighting the effectiveness of the proposed method. Similarly, the BPCER and APCER values show substantial improvements, indicating enhanced robustness to both live and attack samples. Consequently, the ACER decreases from $13.90\%$ to $8.97\%$.

\renewcommand{\arraystretch}{1.2}
\setlength{\tabcolsep}{4pt}

\begin{table*}[htbp]
\centering
\small
\caption{Comparison of the proposed method with ViTranZFAS \cite{george_effectiveness_2021} on the collected dataset.}
\label{tab:collected_dataset_comparison}

\resizebox{\textwidth}{!}{
\begin{tabular}{>{\centering\arraybackslash}m{3.5cm} 
                >{\centering\arraybackslash}m{2cm} 
                >{\centering\arraybackslash}m{2cm} 
                >{\centering\arraybackslash}m{2cm} 
                >{\centering\arraybackslash}m{2cm} 
                >{\centering\arraybackslash}m{2cm}}
\hline
\textbf{Method} & \textbf{Accuracy (\%)} & \textbf{BPCER (\%)} & \textbf{APCER (\%)} & \textbf{ACER (\%)} & \textbf{EER (\%)} \\
\hline
ViTranZFAS \cite{george_effectiveness_2021} & $86.16 \pm 2.38$ & $13.82 \pm 4.20$ & $13.97 \pm 1.27$ & $13.90 \pm 2.26$ & $15.20 \pm 2.19$ \\
\textbf{Proposed Method} & {\boldmath$91.06 \pm 1.30$} & {\boldmath$11.10 \pm 0.92$} & {\boldmath$6.84 \pm 2.16$} & {\boldmath$8.97 \pm 1.20$} & {\boldmath$9.76 \pm 1.12$} \\
\hline
\end{tabular}
}

\end{table*}

Since our primary focus is on detecting disguise makeup attacks as a specific type of spoofing attempt, we adopt Protocol~I, termed \textit{Known Spoof Attack Detection}, of the SiW-Mv2 dataset for evaluation, as it provides a more relevant setup compared to other protocols. This protocol constructs training and testing splits by dividing live subjects and subjects of each spoof pattern into disjoint sets.
Table~\ref{tab:siwmv2_results} summarizes the results in comparison with SRENet~\cite{guo_multi-domain_2022}. As can be observed, our method achieves significant improvements across most attack categories. 
In particular, for the \textit{makeup} attacks, which include \textit{Cosmetics}, \textit{Impersonation}, and \textit{Obfuscation}, the proposed method demonstrates a substantial performance gain. The largest improvement is observed for \textit{Impersonation}, which is conceptually the most similar to disguise makeup attacks, highlighting the effectiveness of our approach in this challenging scenario.
Specifically, the proposed method reduces the ACER for \textit{Impersonation}, \textit{Cosmetics}, \textit{Funny eye}, \textit{Eye mask}, \textit{Mouth mask}, \textit{Paper glass}, \textit{Silicone}, \textit{Paper mask}, and \textit{Print} attacks by $3.6\%$, $1.36\%$, $0.91\%$, $0.91\%$, $0.2\%$, $1.1\%$, $2.71\%$, $0.6\%$, and $0.74\%$, respectively. 
For the \textit{Obfuscation}, \textit{Mask-half}, and \textit{Mannequin} categories, the proposed method preserves a perfect ACER of $0\%$, sustaining flawless detection performance on these attacks. 
The \textit{Transparent} mask category remains particularly challenging, as the attack retains most of the natural facial appearance and closely resembles live samples, thereby reducing the discriminative cues available to the model.
Overall, the results confirm that the proposed architecture not only improves detection in the most relevant category of makeup-related attacks but also delivers competitive or superior performance across a wide range of other 2D and 3D spoofing attacks. 
This suggests that combining localized feature learning from region-specific subnetworks with full-face attention guidance and metric learning contributes significantly to the robustness and generalization of the model across diverse attack scenarios.
\renewcommand{\arraystretch}{1.3} 
\begin{table*}[htbp]
\centering
\caption{Results on the SIW-Mv2 dataset (Known Spoof Attack Detection); 
Obf.: Obfuscation; Imp.: Impersonation; Cos.: Cosmetics; Trans.: Transparent; Man.: Mannequin.}
\label{tab:siwmv2_results}
\begin{tabular}{l *{15}{c}}
\hline
\textbf{Method} & \textbf{Metric} 
& \textbf{Obf.} 
& \textbf{Imp.} 
& \textbf{Cos.} 
& \textbf{\makecell{Funny\\eye}} 
& \textbf{\makecell{Eye\\mask}} 
& \textbf{\makecell{Mouth\\mask}} 
& \textbf{\makecell{Paper\\glass}} 
& \textbf{Mask-half} 
& \textbf{Silicone} 
& \textbf{Trans.} 
& \textbf{\makecell{Paper\\mask}} 
& \textbf{Man.} 
& \textbf{Replay} 
& \textbf{Print} \\
\hline
\multirow{2}{*}{SRENet \cite{guo_multi-domain_2022}} 
 & ACER (\%) & \textbf{0} & 3.6 & 2.7 & 1.1 & 1.1 & 0.2 & 1.1 & \textbf{0} & 5.4 & \textbf{0} & 0.6 & \textbf{0} & \textbf{1.9} & 1.5 \\
 & TDR@FDR=1\% & 100 & 80 & 87.5 & 31.2 & 47.8 & \textbf{100} & 44.8 & \textbf{100} & \textbf{34.3} & \textbf{100} & \textbf{100} & \textbf{100} & 97.4 & \textbf{98.2} \\
\hline
\multirow{2}{*}{Proposed Method} 
 & ACER (\%) & \textbf{0} & \textbf{0} & \textbf{1.34} & \textbf{0.19} & \textbf{0.19} & \textbf{0} & \textbf{0} & \textbf{0} & \textbf{2.69} & 5.38 & \textbf{0} & \textbf{0} & 2.05 & \textbf{0.76} \\
 & TDR@FDR=1\% & \textbf{100} & \textbf{100} & \textbf{91.66} & \textbf{100} & \textbf{100} & \textbf{100} & \textbf{100} & \textbf{100} & 25 & 42.85 & \textbf{100} & \textbf{100} & \textbf{97.43} & 98.18 \\
\hline
\end{tabular}%

\end{table*}

\subsection{Visualization}
\begin{figure*}[t]
    \centering
    \includegraphics[
        width=\textwidth,
        height=0.25\textheight,
    ]{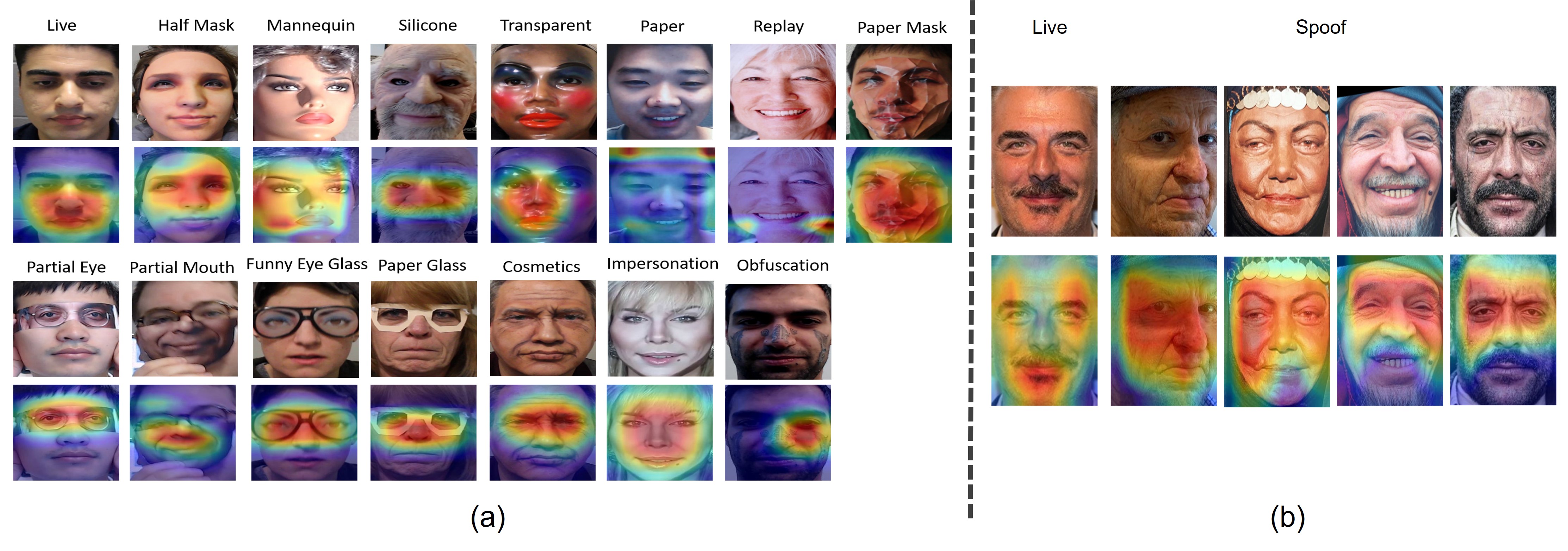}
    \caption{Representative Grad-CAM heatmaps for (a) the SIW-Mv2 dataset and (b) the collected dataset. The highlighted regions correspond to discriminative facial cues leveraged by the model.}
    \label{fig:gradcams}
\end{figure*}

To generate attention score from the Phase 1 and interpret the full-face network, we employ Grad-CAM \cite{selvaraju_grad-cam_2017}. Using gradients from the last convolutional layer, Grad-CAM produces heatmaps that highlight the contribution of local regions to the prediction. Examples are shown in Fig.~\ref{fig:gradcams}, where (a) corresponds to SIW-Mv2 and (b) to the collected dataset; red regions indicate higher importance.

In SIW-Mv2, the heatmaps show that the model captures attack-specific discriminative cues. Live samples produce strong activations in central facial regions (eyes, nose, cheeks, and mouth), while spoof attacks exhibit distinct patterns. Global attacks such as impersonation, mannequin, and mask-based attacks activate broader facial areas, whereas localized attacks (e.g., cosmetics, obfuscation, funny eye, partial eye, paper glasses, and partial mouth) concentrate responses around manipulated regions. Replay and print attacks yield more diffused attention, consistent with global degradations, while transparent masks remain challenging due to their subtle artifacts.

A similar trend is observed in the collected dataset, where activations focus on manipulated regions in disguise samples, while for live faces, the strongest activations appear around the eyes, cheeks, and mouth–chin, which provide stable cues for live detection.

Overall, the heatmaps confirm that the network captures meaningful class-specific patterns, providing interpretability and supporting the patch-based weighting mechanism in the proposed framework.

\section{Discussion}
We show that the proposed attention-guided patch-based framework achieves strong performance on the collected disguise makeup dataset. Experimental results demonstrate that the framework effectively detects makeup presentation attacks and performs well across diverse 2D and 3D spoof types on the SIW-Mv2 dataset \cite{guo_multi-domain_2022}. Furthermore, it achieves competitive performance compared to existing approaches.
In contrast to prior studies that address specific makeup variations under constrained settings or rely on paired reference images of the claimed identity, the proposed framework addresses this challenge without requiring paired images and does not utilize auxiliary cues for presentation attack detection, such as depth information \cite{chen_automatic_2013, alzahrani_deep_2021, rathgeb_detection_2021, guo_multi-domain_2022, zhou_instance-aware_2023}.
Although the framework focuses specifically on detecting a particular type of presentation attack—namely, disguise makeup—the substantial variability in application techniques, materials, and resulting visual appearances within this category necessitates a detection approach that generalizes effectively across diverse samples.
This generalization capability stems from several interrelated design choices. Disguise alterations are inherently local, yet not confined to fixed facial regions, as different application techniques may modify the eyes in one instance and the mouth or cheeks in another. Accordingly, we adopt a patch-based strategy with dedicated subnetworks for seven facial regions, weighted by attention scores derived from a full-face network. As shown in Fig.~\ref{fig:gradcams}, Grad-CAM heatmaps confirm that this attention mechanism effectively captures discriminative cues for both disguised and live faces, highlighting manipulated regions in attack samples and distinctive areas in live samples alike. These attention scores are therefore well suited to guide the patch-based subnetworks toward the most informative regions. Metric learning applied at both full-face and patch levels contributes to improved separation between live and disguised representations, and a whitening transformation ensures that the attention mechanism remains robust to style variations across diverse capture conditions.
Examining the results on the collected dataset, we observe that low image quality—arising from motion blur, poor illumination, or compression artifacts—presents a notable challenge for the model. Since disguise detection relies heavily on fine-grained textural details, degraded inputs tend to obscure these discriminative cues, leading to misclassification, while the relatively small number of such samples in the training set limits the model's robustness under these conditions. Certain subtle attacks, such as transparent masks, also remain challenging due to minimal visual deviation from live faces. Addressing these limitations through improved robustness to severe image degradation and expanded data collection remains an avenue for future work. 
In summary, the combination of region-specific patch analysis, metric learning, and style-invariant attention weighting yields a practical framework with strong generalization for real-world disguise makeup detection.

\section{Conclusion}
The vulnerability of face recognition systems to presentation attacks remains a pressing concern, particularly disguise makeup, which introduces natural-looking modifications through varied techniques and materials. We proposed a generalized framework that integrates a style-invariant full-face network with metric learning to generate region-wise attention via Grad-CAM, guiding a patch-based phase for fine-grained localized analysis. Experimental results demonstrate strong generalization, achieving 8.97\% ACER and 9.76\% EER on the collected disguise dataset, and 0\% ACER on Obfuscation and Impersonation attacks and 1.34\% on Cosmetics attacks within SIW-Mv2, while remaining competitive across other spoof types. In the future, we intend to explore anomaly detection formulations to better model live face distributions, investigate adversarial training strategies to further improve robustness, and extend the framework toward open-set scenarios with unknown attack types.
\bibliographystyle{IEEEtran}
\bibliography{paper_references}



%





\end{document}